\documentclass{article}
\usepackage[preprint]{log_2022}

\usepackage{booktabs}            
\usepackage{multirow}            
\usepackage{amsfonts}            
\usepackage{graphicx}            
\usepackage[numbers,compress,sort]{natbib}
\usepackage{colortbl}
\usepackage{researchpack}
\usepackage[backref=page]{hyperref}
\usepackage[capitalise,nameinlink]{cleveref}
\usepackage{pifont}
\usepackage{enumitem}
\usepackage{caption}
\usepackage{subcaption}

\newcommand{\dm}[1]{#1}

\newcommand{\mbf}[1]{\mathbf{#1}}
\newcommand{\rbf}[1]{\cellcolor{red!60}#1}
\newcommand{\obf}[1]{\cellcolor{orange!40}{#1}}
\newcommand{\ok}{\cellcolor{ForestGreen!40}}

\newcommand{\KL}{\ensuremath{\mathsf{KL}}\xspace}
\newcommand{\UNF}{\ensuremath{\mathsf{Unf}}\xspace}
\newcommand{\FIDP}{\ensuremath{\mathsf{Fid}^{+}}\xspace}
\newcommand{\FIDM}{\ensuremath{\mathsf{Fid}^{-}}\xspace}
\newcommand{\BBBP}{\texttt{BBBP}\xspace}
\newcommand{\MUTAG}{\texttt{MUTAG}\xspace}
\newcommand{\BATWOMOTIF}{\texttt{Ba2Motif}\xspace}
\newcommand{\BAMULTISHAPES}{\texttt{BaMS}\xspace}
\newcommand{\GISST}{\texttt{GISST}\xspace}
\newcommand{\PROTGNN}{\texttt{ProtGNN}\xspace}
\newcommand{\PIGNN}{\texttt{PIGNN}\xspace}
\newcommand{\PIGNNT}{\texttt{PIGNN+T}\xspace}
\newcommand{\PIGNNP}{\texttt{PIGNN+P}\xspace}

\title[How Faithful are Self-Explainable GNNs?]{How Faithful are Self-Explainable GNNs?}
\author[M. Christiansen et al.]{%
    Marc Christiansen\thanks{Equal contribution.}\\
    \institute{CS Department, Aarhus University}\\
    \email{mc@marcs.dk}\\
    \And
    Lea Villadsen\footnotemark[1]\\
    \institute{CS Department, Aarhus University}\\
    \email{lea@villadsen.one}\\
    \AND
    Zhiqiang Zhong\\
    \institute{CS Department, Aarhus University}\\
    \email{zzhong@cs.au.dk}\\
    \AND
    Stefano Teso\\
    \institute{CIMeC \& DISI, University of Trento}\\
    \email{stefano.teso@unitn.it}\\
    \And
    Davide Mottin\\
    \institute{CS Department, Aarhus University}\\
    \email{davide@cs.au.dk}\\
}

\begin{document}
\maketitle

\begin{abstract}
    \textit{Self-explainable} deep neural networks are a recent class of models that can output \textit{ante-hoc} local explanations that are \textit{faithful to the model's reasoning}, and as such represent a step forward toward filling the gap between \textit{expressiveness} and \textit{interpretability}.
    Self-explainable graph neural networks (GNNs) aim at achieving the same in the context of graph data. This begs the question:  \emph{do these models fulfill their implicit guarantees in terms of faithfulness?}
    In this extended abstract, we analyze the faithfulness of several self-explainable GNNs using different measures of faithfulness, identify several limitations -- both in the models themselves and in the evaluation metrics -- and outline possible ways forward.
\end{abstract}

\section{Introduction}

A number of \textit{self-explainable} deep models for image data have been recently developed that sport \textit{high performance} without giving up on \textit{interpretability} \citep{alvarez2018towards, chen2019looks, koh2020concept, chen2020concept, marconato2022glancenets, espinosa2022concept}.
At a high level, they follow a two-level architecture that first maps input images to a set of interpretable, high-level ``concepts'' (using, \eg a convolutional neural network) and then infers predictions based on the extracted concepts only.  This step is implemented with a transparent inference layer (\eg a sparse linear map) from which concept-based explanations can be easily derived.
This makes local explanations \textit{cheap} to compute, \textit{expressive} \citep{kambhampati2022symbols, teso2023leveraging}, and \textit{faithful} to the model's inference process \citep{rudin2019stop}.
See~\citep{schwalbe2022concept} for an overview.

These developments have spurred interest into self-explainable GNN architectures that follow a similar setup and aim at achieving the same feat for graph data.  A brief overview of these architectures is given in \cref{sec:methods}.
This begs the question:  \textit{do self-explainable GNN architectures uphold the promise of producing explanations that are faithful to the model's reasoning?}

In this preliminary investigation, we \textit{focus on graph classification} and provide an initial answer to this question for a representative selection of self-explainable GNN architectures.
Namely, we empirically analyze the faithfulness of explanations output by one information-constrained architecture (\GISST \citep{lin2020graph}) and two prototype-based architectures (\PROTGNN \citep{zhang2022protgnn} and \PIGNN \citep{ragno2022prototype}) on four different data sets, and according to different metrics.

Our results highlight several interesting phenomena.
First, these architectures fail to guarantee faithfulness, thus falling short of the original desideratum.
Furthermore, their explanations exhibit widely different degrees of faithfulness depending on the data set.
Naturally, this hinders trustworthiness.
Second, and equally importantly, we find that well-known measures of faithfulness lack a natural reference, and that therefore can paint an incomplete picture of explanation quality.
Summarizing, our results raise doubts on the interpretability of self-explainable GNN architectures and call for stricter scrutiny of their properties and of the measures proposed to evaluate them.

\section{Post-hoc and Ante-hoc Explanations for Graph Neural Networks}
\label{sec:methods}

\textbf{Explainability and faithfulness.}  Graph Neural Networks (GNNs) are discriminative classifiers that, given an input graph $G = (V, E)$,  annotated with node features $X$, output a conditional distribution over the labels $p_\theta(Y \mid G)$ parameterized by $\theta$.
In this abstract, we focus on \textit{graph classification}, in which case a prediction is readily obtained by solving $\argmax_y p_\theta(y \mid G)$.

The inference process of GNNs relies on opaque operations -- chiefly, message passing of uninterpretable embedding vectors -- and as such it is difficult to interpret for laypeople and experts alike.
Motivated by this, a number of \textit{post-hoc} explainers have been developed for translating the reasoning process of GNNs into human-understandable terms.  Roughly speaking, these algorithms extract an (attributed) \textit{subgraph} $\calE$ of $G$ highlighting those elements -- nodes, edges, features, or some selection thereof -- that are most \textit{responsible} for a GNN's prediction.
See \citep{longa2022explaining, agarwal2023evaluating, kakkad2023survey} for an overview.

A key desideratum is that of \textit{faithfulness}, that is, explanations should portray a faithful picture of the model's decision making process \citep{yuan2022explainability, amara2022graphframex, agarwal2023evaluating}.  Unless explanations are faithful, they can fool users into, \eg wrongly trusting misbehaving models, and can complicate debugging \citep{teso2023leveraging}.
Metrics for evaluating faithfulness will be discussed in \cref{sec:experiments}.
Post-hoc approaches give limited guarantees about faithfulness \citep{longa2022explaining, agarwal2023evaluating}, prompting the development of \textit{self-explainable} GNNs able to produce explanations that are -- at least on paper -- faithful by design, see \citep[Section~3.2]{kakkad2023survey}.

\textbf{Self-explainability via information constraints.}  A first group of approaches encourage interpretability by ensuring the information used for prediction crosses an \textit{information bottleneck}, oftentimes implemented as a sparse subgraph, and read off an explanation from this \citep{miao2022interpretable,miao2022interpretablerandom,yu2020graph,yu2022improving}.
We focus on a representative of this group, \GISST \citep{lin2020graph}, a model-agnostic approach for turning any GNN into an information-constrained self-explainable model.
In short, \GISST introduces an attention layer on the model embeddings and encourages this layer -- via an appropriate sparsification penalty over the features and the edges of the graph -- to focus on a subgraph $\calE$ of $G$ (and a subset of features $\calX$) capturing all information necessary to infer a label. The method computes an importance score for each edge in the graph. The explanation subgraph is the set of edges with the highest importance that connect a predefined number of nodes.

\textbf{Self-explainability via structure constraints.}  Another class of self-explainable GNNs seek interpretability by imposing some form of architectural bias \citep{dai2021towards, dai2022towards, feng2022kergnns}.
A popular strategy involves augmenting GNNs with prototypes.  These are (embeddings of) concrete training examples or parts thereof, and can be used to implement case-based reasoning \citep{kim2014bayesian}.
One example is \PROTGNN \citep{zhang2022protgnn}, which first embeds inputs using a GCN~\citep{kipf2016semi} or a GAT~\citep{velivckovic2017graph}, and then infers a prediction by matching the input graph with learned prototypes.  In turn, these prototypes are learned so as to activate highly on training examples corresponding to a specific target class.
Explanations are then encoded in terms of what parts of the input graph have triggered the activation of the learned prototypes.

\PIGNN \citep{ragno2022prototype} is similar, in that it infers concept activations based on the similarity between the input graph and learned prototypes.  The similarity is however computed at the level of individual node embeddings.  Node predictions are then computed directly from the prototype activations, while graph predictions make use of a max pooling operator.
\PIGNN depends on an underlying prototype-based neural network.  This can be either ProtoPNet~\citep{chen2019looks} (yielding \PIGNNP) or TesNet~\cite{wang2021interpretable} (yielding \PIGNNT).
TesNet is a prototype architecture for images that encourages disentanglement between concepts by enforcing orthonormality of the subspaces spanned by the concepts.
Both models return the subgraphs that activate the prototypes associated to a given class the most by adapting existing image-based ProtoPNets by substituting convolutional layers with a GCN and adding a max-pooling before the classification layer to aggregate node-prototypes into subgraphs.

\section{Empirical Analysis}
\label{sec:experiments}

We address the following research questions:
\textbf{Q1}: Do self-explainable GNNs guarantee faithfulness in practice?
\textbf{Q2}: Do existing faithfulness metrics properly gauge the quality of the model's explanations?
To this end, we integrated the implementations of \GISST, \PROTGNN and \PIGNN from the original papers into a unified framework built on Pytorch Geometric \citep{pyg}. We publish the code and the data for reproducibility at \url{https://anonymous.4open.science/r/SEGNNEval}.

\textbf{Data sets}.  We consider four graph classification data sets, cf.~\cref{table:datasets}.
\BBBP \citep{wu2018moleculenet} and \MUTAG \citep{debnath1991structure, morris2020tudataset} are data sets of chemical compounds labeled according to their blood-brain barrier penetration ability and mutagenicity, respectively.
\BATWOMOTIF \cite{luo2020parameterized} and BAMultiShapes (\BAMULTISHAPES for short) \citep{azzolin2022global} are synthetic data sets.  Similarly to BA-Shapes \citep{ying2019gnnexplainer}, graphs are obtained by linking a random Barabasi-Albert (BA) graph \citep{albert2002statistical} to one or more motifs that cannot be generated by the BA process.  In \BATWOMOTIF, the motif is either a house or a $5$-node cycle, and it fully determines the label.  In \BAMULTISHAPES, it is either a house, a wheel or a grid, and the label is determined by the presence of exactly two out of three motifs.  For the BA data sets, node features encode the number of edges and triangles the node participates in, and the BA subgraphs were generated using a fixed seed, for reproducibility.

\begin{table}[!t]
    \caption{Overview of data sets used in our experiments.  Numbers are averaged.}
    \label{table:datasets}
    \centering
    \scriptsize
    \begin{tabular}{lrrrrr} 
        \toprule
        \textsc{Dataset}
            & \textsc{\# Graphs}
            & \textsc{Avg. \# Nodes}
            & \textsc{Avg. \# Edges}
            & \textsc{\# Features}
            & \textsc{\# Classes}
        \\
        \midrule
        \MUTAG
            & $188$
            & $17.9$
            & $39.6$
            & $7$
            & $2$
        \\
        \BBBP
            & $2039$
            & $24.0$
            & $51.9$
            & $9$
            & $2$
        \\
        \BATWOMOTIF
            & $1000$
            & $30.0$
            & $57.8$
            & $2$
            & $2$
        \\
        \BAMULTISHAPES
            & $1000$
            & $40.0$
            & $87.4$
            & $2$
            & $2$
        \\
        \bottomrule
    \end{tabular}
\end{table}

\textbf{Faithfulness metrics}.  We evaluate faithfulness using three metrics.
Let $\calE$ be the explanation subgraph induced by the nodes, edges, and features deemed \textit{relevant} by the model, and $\calC$ be its complement.

\textit{Unfaithfulness}~\citep{agarwal2023evaluating} measures how strongly a prediction depends on \textit{irrelevant} nodes, edges, and features:
\[
    \UNF
        = 1 - \exp \big(
            -\KL \big(
                p_\theta(Y \mid G)
                \ \| \ 
                p_\theta(Y \mid \calE)
            \big)
        \big)
    \label{eq:unf}
\]
Unfaithfulness is close to zero only if the Kullback-Leibler divergence $\KL$ is small, that is, if $G$ and $\calE$ yield similar label distributions\footnote{In \cite{agarwal2023evaluating}, $\calE$ is obtained by zeroing-out the irrelevant features from $G$.  In this abstract, $\calE$ is obtained by also deleting irrelevant nodes and edges, so as to take the relevance of topological elements into account.}.

\textit{Fidelity}~\citep{yuan2022explainability, amara2022graphframex} includes two metrics assessing to what degree explanations are \textit{necessary} ($\FIDP$) and \textit{sufficient} ($\FIDM$).  These are defined as\footnote{Our definition pertains to graph classification.  Fidelity is also defined for node classification~\citep{yuan2022explainability}.}:
\[
    \FIDP
        = \big|
            \Ind{\widehat{y} = y} - \Ind{\widehat{y}^\calC = y}
        \big|,
    \quad
    \FIDM
        = \big|
            \Ind{\widehat{y} = y} - \Ind{\widehat{y}^\calE = y}
        \big|
    \label{eq:fid}
\]
Here, $\widehat{y}^\calE = \argmax_y \ p_\theta(y \mid \calE)$ and similarly for $\widehat{y}^\calC$.
High \FIDP means that $\calE$ is \textit{necessary}, in the sense that the prediction does change when the model is fed the irrelevant subgraph $\calC$ only, while low \FIDM means the explanation is \textit{sufficient}, in that the prediction remains unchanged when considering the relevant subgraph $\calE$ only.

These measures differ in that \FIDP and \FIDM look at the hard predictions and depend on the ground-truth label, whereas \UNF considers the change in label distribution only.  As such, it can be viewed as a smoother version of \FIDM and behaves similarly in our experiments.

\textbf{Q1:  Faithfulness of self-explainable GNNs is not perfect and varies across tasks}. \cref{table:results} shows the average result of $5$ runs of the algorithm with 80/10/10 train, validation, and test split.
Most architectures attained good ($>80\%$) and stable ($<10\%$ std. dev.) graph classification accuracy on all data sets.  The exception is \PROTGNN, which fared very poorly on the synthetic tasks.  Since it tends to over-predict the majority class (especially in \BATWOMOTIF), faithfulness measures are unreliable.  Hence, we excluded \PROTGNN from our analysis for these data sets.

In stark contrast, their faithfulness --indicated as $\UNF(\calE)$, $\FIDP(\calE)$, and $\FIDM(\calE)$ -- is far from perfect.
All models tend to perform better at \UNF and \FIDM (best results are $0$ and $0.07$, respectively, the lower the better) than at \FIDP (best result is $0.58$, the higher the better), indicating they are better suited at generating \textit{necessary}, rather than \textit{sufficient}, explanations. Note that in the case of \PROTGNN on \MUTAG, \UNF might be zero due to negligible changes in the prediction vector and rounding in the results, while \FIDM, although being very low, is $0.1$ indicating that some of the predictions swaps classes while the overall distribution is stable. This example shows how \FIDM and and \UNF captures the same phenomenon but with different granularity. 

Moreover, explanation faithfulness varies widely across learning tasks.
For instance, \PIGNNP attains excellent \UNF and \FIDM in \BBBP ($0.07$ for both), but fares poorly in \BATWOMOTIF ($0.50$ and $0.59$, respectively).  A $10$-$15\%$ difference in unfaithfulness also exists for the other models.
As for \FIDP, the delta across data sets is $34\%$ for \GISST, $0.53$ for \PIGNNP, $0.18$ for \PIGNNT, and $0.13$ for \PROTGNN.
Summarizing, no model yields consistent faithfulness across tasks for any of the metrics.  This makes it difficult to fully trust the explanations output by these models and raises doubts on their applicability to high-stakes tasks.

\textbf{Q2: Absolute faithfulness measures can be misleading}.  It is true, however, that faithfulness is excellent for some combinations of models and tasks.  This is especially the case for \PROTGNN on \MUTAG and for \PIGNNT on \BATWOMOTIF, which both attain near-perfect \UNF and \FIDM.
Here, we argue that the absolute values in faithfulness can lead to erroneous interpretations.   In a nutshell, faithfulness measures require a baseline method to establish the improvement.
We offer such a term of reference by compare each measure by the same measure computed on a random explanation model that generates subgraphs $\calR$ of $G$ of the same size of the explanation $\calE$.  That is, we replace $\calE$ with $\calR$ in \cref{eq:unf} and \cref{eq:fid}.
We indicate the results as $\UNF(\calR)$, $\FIDM(\calR)$ and $\FIDP(\calR)$ in Table~\ref{table:results}.

We can see that oftentimes the model's explanations are in fact more faithful than the random ones, especially for \MUTAG data.  Yet, in many cases the models produce worse or comparable explanations to a random graph. Some results are especially interesting.  For instance, \PROTGNN is the most faithful method in \MUTAG, while our analysis displays the same behavior on random explanations. Another interesting case is that of \GISST, which performs the best on \BAMULTISHAPES while exhibiting \UNF $45\%$ worse than random.
In conclusion, while faithfulness measures offer an intuitive idea of the model's explanations, they provide only a partial view of their overall quality.

\begin{table}[!t]
    \centering
    \caption{Results for all models and data sets.  We report average and std. dev. of classification accuracy, faithfulness of the model explanations $\calE$, and faithfulness of random subgraphs \dm{$\calR$}.
    \textbf{Bold} entries indicate best result, \textbf{\textcolor{red}{red}} ones that the random subgraph is at least as faithful as the model's explanation, \textbf{\textcolor{orange}{orange}} ones that it is \dm{at most} $0.10$ worse, and \textbf{\textcolor{ForestGreen}{green}} the rest.}
    \label{table:results}

    \setlength{\tabcolsep}{3pt}

    \scriptsize
    \begin{tabular}{ll|c|cc|cc|cc}
        \toprule
        \textsc{Dataset}
            & \textsc{Model}
            & \textsc{Acc} $(\uparrow)$
            & $\UNF(\calE) (\downarrow)$
            & $\UNF(\calR) (\downarrow)$
            & $\FIDM(\calE) (\downarrow)$
            & $\FIDM(\calR) (\downarrow)$
            & $\FIDP(\calE) (\uparrow)$
            & $\FIDP(\calR) (\uparrow)$
        \\
        \midrule
        \multirow{4}{*}{\BBBP}
        & \GISST        & $\mbf{0.87 \pm 0.01}$
                        & $0.22 \pm 0.15$ & $\rbf{0.15 \pm 0.02}$
                        & $0.29 \pm 0.25$ & $\rbf{0.21 \pm 0.04}$
                        & $0.16 \pm 0.01$ & $\rbf{0.20 \pm 0.03}$ \\
        & \PIGNNP       & $0.86 \pm 0.01$
                        & $\mbf{0.07 \pm 0.02}$ & $\obf{0.16 \pm 0.03}$
                        & $\mbf{0.07 \pm 0.03}$ & \ok $0.20 \pm 0.03$
                        & $\mbf{0.33 \pm 0.06}$ & \ok $0.22 \pm 0.03$ \\
        & \PIGNNT       & $0.86 \pm 0.01$
                        & $0.10 \pm 0.01$ & $\obf{0.15 \pm 0.01}$
                        & $0.14 \pm 0.02$ & \ok $0.22 \pm 0.02$
                        & $0.30 \pm 0.05$ & $\obf{0.27 \pm 0.03}$ \\
        & \PROTGNN      & $0.85 \pm 0.01$
                        & $0.15 \pm 0.08$ & \ok $0.27 \pm 0.05$
                        & $0.14 \pm 0.02$ & $\obf{0.23 \pm 0.05}$
                        & $0.21 \pm 0.06$ & $\obf{0.20 \pm 0.05}$ \\
        \midrule
        \multirow{4}{*}{\MUTAG}
        & \GISST        & $\mbf{0.86 \pm 0.04}$
                        & $0.11 \pm 0.12$ & $\obf{0.13 \pm 0.06}$
                        & $0.31 \pm 0.17$ & \ok $0.41 \pm 0.16$
                        & $\mbf{0.50 \pm 0.19}$ & \ok $0.38 \pm 0.13$ \\
        & \PIGNNP       & $0.82 \pm 0.02$
                        & $0.12 \pm 0.08$ & $\obf{0.18 \pm 0.08}$
                        & $0.26 \pm 0.26$ & \ok $0.59 \pm 0.08$
                        & $0.42 \pm 0.17$ & $\rbf{0.52 \pm 0.22}$ \\
        & \PIGNNT       & $0.84 \pm 0.07$
                        & $0.14 \pm 0.13$ & $\rbf{0.08 \pm 0.07}$
                        & $0.40 \pm 0.30$ & \ok $0.51 \pm 0.12$
                        & $0.37 \pm 0.05$ & $\rbf{0.40 \pm 0.15}$ \\
        & \PROTGNN      & $0.82 \pm 0.02$
                        & $\mbf{0.00 \pm 0.00}$ & $\rbf{0.00 \pm 0.00}$
                        & $\mbf{0.10 \pm 0.07}$ & $\obf{0.18 \pm 0.04}$
                        & $0.34 \pm 0.05$ & \ok $0.20 \pm 0.05$ \\
        \midrule
        \multirow{4}{*}{\BATWOMOTIF}
        & \GISST        & $0.98 \pm 0.02$
                        & $0.13 \pm 0.16$ & \ok $0.32 \pm 0.09$
                        & $\mbf{0.14 \pm 0.16}$ & \ok $0.32 \pm 0.08$
                        & $0.51 \pm 0.02$ & $\rbf{0.52 \pm 0.02}$ \\
        & \PIGNNP       & $\mbf{1.00 \pm 0.00}$
                        & $0.50 \pm 0.04$ & $\rbf{0.42 \pm 0.01}$
                        & $0.59 \pm 0.06$ & $\rbf{0.50 \pm 0.00}$
                        & $0.01 \pm 0.01$ & $\rbf{0.50 \pm 0.00}$ \\
        & \PIGNNT       & $0.98 \pm 0.01$
                        & $\mbf{0.03} \pm 0.02$ & $\obf{0.08 \pm 0.05}$
                        & $0.21 \pm 0.06$ & \ok $0.57 \pm 0.06$
                        & $\mbf{0.58 \pm 0.14}$ & $\obf{0.56 \pm 0.08}$ \\
        & \PROTGNN      & $0.51 \pm 0.11$
                        & -- & --
                        & -- & --
                        & -- & -- \\
        \midrule
        \multirow{4}{*}{\BAMULTISHAPES}
        & \GISST        & $0.79 \pm 0.13$
                        & $\mbf{0.11} \pm 0.07$ & $\obf{0.13 \pm 0.06}$
                        & $\mbf{0.20 \pm 0.13}$ & $\obf{0.27 \pm 0.12}$
                        & $0.42 \pm 0.20$ & $\obf{0.39 \pm 0.19}$ \\
        & \PIGNNP       & $\mbf{0.96 \pm 0.00}$
                        & $0.32 \pm 0.00$ & $\rbf{0.32 \pm 0.00}$
                        & $0.54 \pm 0.00$ & $\rbf{0.54 \pm 0.00}$
                        & $0.54 \pm 0.00$ & $\rbf{0.54 \pm 0.00}$ \\
        & \PIGNNT       & $0.92 \pm 0.00$
                        & $0.13 \pm 0.01$ & \ok $0.30 \pm 0.00$
                        & $0.46 \pm 0.01$ & $\rbf{0.56 \pm 0.00}$
                        & $\mbf{0.56 \pm 0.00}$ & $\rbf{0.56 \pm 0.00}$ \\
        & \PROTGNN      & $0.64 \pm 0.10$
                        & -- & --
                        & -- & --
                        & -- & -- \\
        \bottomrule
    \end{tabular}
\end{table}

\section{Discussion and Conclusion}
\label{sec:discussion-conclusion}

Our initial results indicate that, at least when it comes to graph classification, it is \textit{hard to assert that self-explainable GNNs are faithful by design}.  Specifically, faithfulness of these models seems to be very data set dependent.  This becomes even more clear when we compare the faithfulness of their explanations to that of a completely uninformed baseline.

One possible motivation behind these results is that the explanations offered by self-explainable models are faithful \textit{at the concept level}.  For instance, take prototype-based architectures.  These extract explanations from an interpretable top layer feeding on the prototypes themselves, hence these explanations are faithful \textit{by construction} to the prototypes.  In order to convert this into a subgraph, they have to ground the explanation in terms of nodes, edges, and features, rather than prototypes.  This mapping is usually inexact (or, more specifically, not bijective), and as such the resulting low-level explanation no longer has faithfulness guarantees.  A more detailed analysis of the root causes of unfaithfulness for other self-explainable architectures is however left to future work.

Naturally, our results are preliminary.  In follow-up work, we plan to verify to what extent they carry over to other self-explainable GNN architectures and to extend our analysis to \textit{node classification}.
Another major issue is whether the \textit{concepts} (\eg prototypes) learned by these models are in fact guaranteed to be interpretable.  Yet, this issue is shared by self-explainable models for image data as well, and needs a separate investigation to be understood properly \citep{marconato2022glancenets}.

\section*{Acknowledgements}
The research of ST was partially supported by TAILOR, a project funded by EU Horizon 2020 research and innovation programme under GA No 952215.
Davide Mottin was partially supported by the Horizon Europe and Danmarks Innovationsfond under the Eureka, Eurostar grant no E115712.

\bibliographystyle{unsrtnat}
\bibliography{main, explanatory-supervision}
\end{document}